\documentclass[preprint, 12pt]{elsarticle}

\usepackage{epsfig}
\usepackage{graphicx}
\usepackage{amsmath}
\usepackage{amsthm}
\usepackage{amsbsy}
\usepackage{amsfonts}
\usepackage{subfigure}
\usepackage{float}
\usepackage{url}
\usepackage{tabularx}
\usepackage{makecell}
\usepackage{dsfont}
\usepackage{threeparttable}
\usepackage{color}
\usepackage[justification=centering]{caption}
\usepackage{epstopdf}
\usepackage[lined,boxed,commentsnumbered, ruled]{algorithm2e}
\usepackage[normalem]{ulem}
\usepackage{verbatim}
\usepackage{multicol,multirow}
\graphicspath{{./figs/}}

\begin{document}
\begin{frontmatter}
\title{Shared Generative Latent Representation Learning for Multi-view Clustering}

\author[label1]{Ming Yin}
\author[label1]{Weitian Huang}
\author[label2]{Junbin Gao}
\address[label1]{School of Automation, Guangdong University of Technology, Guangzhou, China.}
\address[label2]{The University of Sydney Business School, The University of Sydney, Camperdown, NSW 2006, Australia. \\ e-mail:yiming@gdut.edu.cn, 583501947@qq.com, junbin.gao@sydney.edu.au.}
\begin{abstract}
Clustering multi-view data has been a fundamental research topic in the computer vision community. It has been shown that a better accuracy can be achieved by integrating information of all the views than just using one
view individually. However, the existing methods often struggle with the issues of dealing with the large-scale datasets and the poor performance in reconstructing samples. This paper proposes a novel multi-view clustering method by learning a shared generative latent representation that obeys a mixture of Gaussian distributions. The motivation is based on the fact that the multi-view data share a common latent embedding despite the diversity among the views. Specifically, benefited from the success of the deep generative learning, the proposed model can not only extract the nonlinear features from the views, but render a powerful ability in capturing  the correlations among all the views. The extensive experimental results, on several datasets with different scales, demonstrate that the proposed method outperforms the state-of-the-art methods under a range of performance criteria.
\end{abstract}
\begin{keyword}
Multi-view clustering, Representation learning, Generative model, Mixture of Gaussian, Large-scale clustering
\end{keyword}
\end{frontmatter}
\section{Introduction}
Image clustering is one of the fundamental research topics, which has been widely studied in computer vision and machine learning \cite{CaronBojanowskiJoulinDouze2018,ChangWangMengXiangPan2017}. As well, as a class of unsupervised learning methods, clustering has attracted significant attention from various applications. With the advance of information technology, in many real-world scenarios, many heterogeneous visual features, such as HOG \cite{DalalTriggs2005}, SIFT \cite{DengDongSocherLiLi2009} and LBP \cite{OjalaPietikainenMaenpaa2002} can be readily acquired and form a new type data, i.e., multi-view data. Therefore, to efficiently capture the complementary information among different views, multi-view clustering has gained considerable attention in the recent years for learning more comprehensive information \cite{Sun2013,XuTaoXu2013}. In essence, multi-view clustering seeks to partition data points based on multiple representations by assuming that the same cluster structure is shared across all the views \cite{GaoNieLiHuang2015,WangNieHuang2013,YinGaoXieGuo2018}. It is crucial for learning algorithm to incorporate the heterogeneous view information to enhance its accuracy and robustness.

In general, multi-view clustering can be roughly separated into two classes, i.e., similarity-based and feature-based. The former aims to construct an affinity matrix whose elements define the similarity between each pair of samples. In light of this, multi-view subspace clustering is one of the most famous similarity-based methods, which purses a latent subspace shared by multiple views, assuming that each views is built from a common subspace \cite{ChaudhuriKakadeLivescuSridharan2009,GaoNieLiHuang2015,YinGaoXieGuo2018,ZhangFuLiuLiuCao2015}. However, these methods often suffer scalability issue due to super-quadratic running time for computing spectra \cite{JiangZhengTanTangZhou2017}. While for the feature-based methods, it seeks to partition the samples into $K$ clusters so as to minimize the within-cluster sum of squared errors, such as multi-view $K$-means clustering \cite{CaiNieHuang2013,XuHanNieLi2017}. It is clear that the selection of feature space is vital as the clustering with Euclidean distance on raw pixels is completely ineffective.

Inspired by the recent amazing success of deep learning in feature learning \cite{HintonSalakhutdinov2006}, a surge of multi-view learning based on deep neural networks (DNN) are proposed \cite{NgiamKhoslaKimNamLeeNg2011,WangAroraLivescuBilmes2015,XuGuanZhaoNiuWangWang2018}. First, Ngiam \emph{et al.} \cite{NgiamKhoslaKimNamLeeNg2011} explored extracting shared representations by training a bimodal deep autoencoders. Next, by extending canonical correlation analysis (CCA), Wang \emph{et al.} \cite{WangAroraLivescuBilmes2015} proposed a novel deep canonically correlation autoencoders (DCCAE), which introduces an autoencoder regularization term into deep CCA. However, unfortunately the aforementioned can only be feasible to the two-view case, 
failing to handle the multi-view one. To explicitly summarize the consensus and complementary information in multi-view data,  a Deep Multi-view Concept learning (DMCL) \cite{XuGuanZhaoNiuWangWang2018} is presented by performing non-negative factorization on every view hierarchically.

Though these methods perform well in multi-view clustering, the generative process of multi-view  data cannot be modeled such that they 
can be used to generate samples accordingly. To this end, benefited from the success of approximate Bayesian inference, the variational autoencoders (VAE) has been the most popular algorithm under the framework that combines differentiable models with variational inference \cite{KingmaWelling2014,PuGanHenaoYuanLiStevensCarin2016}. By modeling the data generative procedure with a Gaussian Mixture Model (GMM) model and a neural network, Jiang \emph{et al.} \cite{JiangZhengTanTangZhou2017} proposed a novel unsupervised generative clustering approach within the framework of VAE, namely Variational Deep Embedding (VaDE).
Although it has shown great advantages in clustering, it is not able to be applied {\it directly} to multi-view learning.

Targeting for classification and information retrieval, Srivastava \emph{et al.} \cite{SrivastavaSalakhutdinov2014} presented a deep Boltzmann machine for learning a generative model of multi-view data. However, until recently there was no successful multi-view extension to clustering yet. The main obstacle is how to efficiently exploit the shared generative latent representation across the views in {\it unsupervised} way. To tackle this issue, in this paper, we propose a novel multi-view clustering by learning a shared generative latent representation that obeys a mixture of Gaussian distributions, namely Deep Multi-View Clustering via Variational Autoencoders (DMVCVAE). In particular, our motivation is based on the fact that the multi-view data share a common latent embedding despite the diversity among the views. Meanwhile, the proposed model  benefits from the success of the deep generative learning, which can capture the data distribution by neural networks.

In summary, our contributions are as follows.
\begin{itemize}
  \item We present to learn a shared generative latent representation for multi-view clustering. Specifically, the generative approach assumes that the data of different views share a commonly conditional distribution of hidden variables given observed data and the hidden data are sampled independently from a mixture of Gaussian distributions.
 \item To better exploit the information from multiple views, we introduce a set of non-negative
 combination weights
 which will be learned jointly with the deep autoencoders network in a unified framework.
 \item We conduct a number of numerical experiments
 showing that the proposed method outperforms the state-of-the-art clustering models on several famous datasets including large-scale multi-view data.
\end{itemize}
\section{Related Works}
In literature, there are a few studies on clustering using deep neural networks \cite{JiZhangLiSalzmannReid2017,PengXiaoFengYauYi2016,TianGaoCuiChenLiu2014,XieGirshickFarhadi2016,YangFuSidiropoulosHong2017}. In a sense, the algorithms are roughly divided into two categories, i.e., separately and jointly deep clustering approaches. The earlier deep clustering algorithms \cite{JiZhangLiSalzmannReid2017,PengXiaoFengYauYi2016,TianGaoCuiChenLiu2014} often work in two stages: firstly, extracting deep features 
and performing traditional clustering successively, such as the $K$-means and spectral clustering, for the final segmentation. Yet the separated process does not help learn clustering favourable features. 
To this end, the jointly feature 
learning and clustering methods \cite{XieGirshickFarhadi2016,YangFuSidiropoulosHong2017} are proposed based on deep neural networks. In 
\cite{XieGirshickFarhadi2016}, Xie \emph{et al.} presented Deep Embedded Clustering (DEC) to learn a mapping from the data space to a lower-dimensional feature space, where it iteratively optimizes a Kullback-Leibler (KL) divergence based clustering objective. In 
\cite{YangFuSidiropoulosHong2017}, Yang \emph{et al.} proposed a dimensionality reduction jointly with $K$-means clustering framework, where deep neural networks are applied to dimensionality reduction.

However, due to the limitation of the similarity measures in the aforementioned methods, the hidden, hierarchical dependencies in the latent space of data are often not able to be captured effectively. Instead, deep generative models were built to better handle the rich latent structures within data \cite{JiangZhengTanTangZhou2017}. In essence, deep generative models are utilized to estimate the density of observed data under some assumptions about its latent structure, i.e., the hidden causes. Recently, Jiang \emph{et al.} \cite{JiangZhengTanTangZhou2017} 
proposed a novel clustering framework, by integrating VAE and a GMM for clustering tasks, namely Variational Deep Embedding (VaDE). Unfortunately, as this method mainly focuses on single-view data, the complementary information from multiple heterogeneous views cannot be efficiently exploited. In other words, the existing generative model cannot deal with the shared latent representations for modeling the generative process of each view data.
\section{The Proposed Method}
\subsection{The Architecture}
Given a collection of multi-view data set $\{\mathrm X^{(v)} \in \mathbb{R}^{d_v \times n} \}$ ($v = 1, 2, ..., m$), totally $m$ views, it is reasonable to assume that the $i$-th sample of the $v$-th view $\mathrm x_i^{(v)} \in \mathbb{R}^{d_v}$ is generated by some 
unknown process, for example, from an unobserved continuous variable $\rm z\in\mathbb{R}^d$. The variable $\rm z$ is a common hidden representation shared by all views. Furthermore, in a typical setting, each sample $\mathrm x^{(v)}$ of a view is assumed to be generated through a two-stage process:
first the hidden variable $\rm z$ is generated according to some prior distribution and then the observed sample $\mathrm x^{(v)}$ is yielded by some conditional distributions $p_{\theta^{(v)}}(\mathrm x^{(v)} | \rm z)$. Usually, due to the unknown of the $\rm z$ and parameters $\theta$, the prior $p_{\theta}(\mathrm{z})$ and the likelihood  $p_{\theta^{(v)}}(\mathrm{x}^{(v)}|\rm z)$ are hidden.

\begin{figure*}
\centering
\includegraphics[width=12cm]{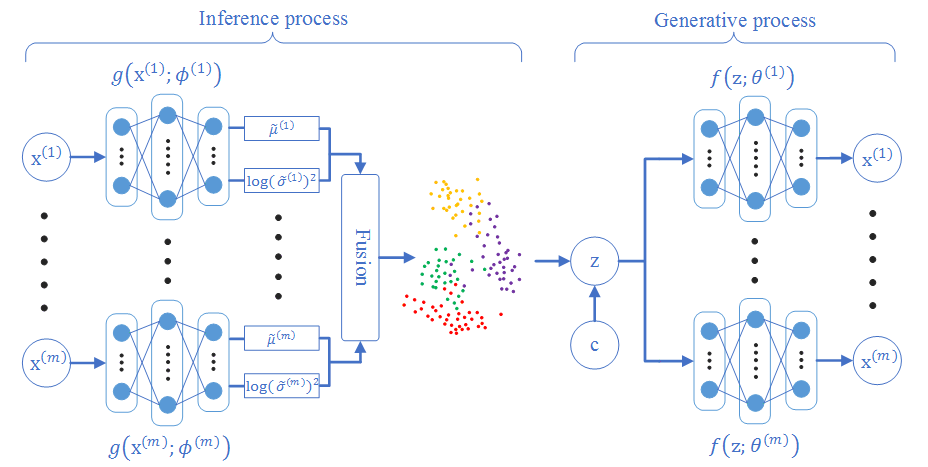}
\caption{The architecture of the proposed multi-view model. The data generative process under the deep autoencoders framework is performed in three steps. (a). A cluster is first picked from a pretrained GMM model; (b). A shared latent representation (embedding) weighted by each view is generated by the prior picked cluster; (c) DNN $f(\mathrm z; \theta^{(v)})$ decodes the latent embedding into an observable $\rm x$. To optimize the ELBO of the proposed model, the encoder network $g(\cdot)$ is applied. }
\label{fig:model}
\end{figure*}

For clustering tasks, it is desired that the observed sample is generated jointly according to the latent variable $\mathrm z$ and an assumed clustering variable $c$.However, the most existing variational autoencoders are not suitable for clustering tasks by design, even to say nothing of multi-view clustering.  Therefore, we are motivated to present a novel multi-view clustering under the VAE framework, by incorporating clustering-promoting objective intuitively. Ideally we shall assume that the sample generative process is given by the new likelihood  $p_{\theta^{(v)}}(\mathrm{x}^{(v)}|\rm z, c)$, conditioned on both the hidden variable $\rm z$ and the cluster label $c$. However for simplicity we break the direct dependence of $\mathrm{x}^{(v)}$ on $c$ conditioned on an assumed Gaussian mixture variable $\rm z$. The proposed framework is shown in the right panel of Figure \ref{fig:model}. In this architecture, multi-view samples $\{\mathrm{x}^{(v)}\}$ are generated by using DNN $f(\cdot)$ to decode the common hidden variable $\mathrm{z}$, which is sampled by GMM as we assumed. To efficiently infer the posterior of both $\rm z$ and $c$ from the information of multiple views, a novel weighted target distribution is introduced, based on individual variational distribution of $\rm z$ from each view. In order to optimize the evidence lower bound (ELBO), similar to VAE, we use DNN $g(\cdot)$ to encode observed data and incorporate the distribution of multiple embeddings to infer the shared latent representation $\mathrm{z}$.

\subsection{The Objective}
For the sake of simplicity, we express a generic multi-view variable as $\{\mathrm{x}^{(v)}\}  := \{\mathrm{x}^{(1)}, ..., \mathrm{x}^{(v)}, ..., \mathrm{x}^{(m)}\}$ where $\mathrm{x}^{(v)}$ is the general variable of the $v$-th view. Consider the latent variables $\mathrm{z}$ and the discrete latent variable $c$ ($c=1, 2, \cdots, K$).
Without loss of generality, in light of clustering task under the framework of VAE, we aim to compute the probabilistic cluster assignments of $\{\mathrm{x}^{(v)}\}$ for each view, denoted by $p(\mathrm{z},c|\{\mathrm{x}^{(v)}\})$. By the Bayes theorem, the corresponding posterior of $\rm z$ and $c$ given $\{\mathrm{x}^{(v)}\}$  is computed as follow.
\begin{equation}
p(\mathrm{z},c|\{\mathrm{x}^{(v)}\})=\frac{p( \{\mathrm{x}^{(v)}\}|\mathrm{z},c)p(\mathrm{z},c)}{\int_{\mathrm{z}}\sum_{c}p(\{\mathrm{x}^{(v)}\}|\mathrm{z},c)p(\mathrm{z},c)d\mathrm{z}} ,
\label{eqn:px_cz}
\end{equation}
where we assume the views are independent, i.e., $p(\{\mathrm{x}^{(v)}\}|\mathrm{z},c) = \prod^m_{v=1}p(\mathrm{x}^{(v)}|\mathrm{z},c)$\footnote{Hereafter the model parameter $\theta^{(v)}$ is omitted.}.

As the integral is intractable, it is hard to calculate the posterior. Inspired by the principle of VAE  \cite{KingmaWelling2014}, we turn to compute an appropriate posterior $q(\mathrm{z},c|\{\mathrm{x}^{(v)}\})$ to approximate the true posterior $p(\mathrm{z},c|\{\mathrm{x}^{(v)}\})$ by minimizing the following KL divergence between them.
\begin{flalign}
&D_{KL}(q(\mathrm{z},c| \{\mathrm{x}^{(v)}\})||p(\mathrm{z},c| \{\mathrm{x}^{(v)}\}) \notag \\
&= \int_{\mathrm{z}}\sum_{c}q(\mathrm{z},c|\{\mathrm{x}^{(v)}\})\log\frac{q(\mathrm{z},c|\{\mathrm{x}^{(v)}\})}{p(\mathrm{z},c| \{\mathrm{x}^{(v)}\})}d\rm{z}\nonumber\\
&= -E_{q(\mathrm{z},c|\{\mathrm{x}^{(v)}\})}\left[\log\frac{p( \{\mathrm{x}^{(v)}\},\mathrm{z},c)}{q(\mathrm{z},c| \{\mathrm{x}^{(v)}\})}\right]+\log p(\{\mathrm{x}^{(v)}\}),
 \label{eqn:kl}
\end{flalign}
where
\begin{eqnarray}
\mathcal{L}_{\mathrm{ELBO}}(\{\mathrm{x}^{(v)}\})\overset{\triangle}{=}E_{q(\mathrm{z},c|\{\mathrm{x}^{(v)}\})}\left[\log\frac{p(\{\mathrm{x}^{(v)}\},\mathrm{z},c)}{q(\mathrm{z},c|\{\mathrm{x}^{(v)}\})}\right]
\label{eqn:elbo}
\end{eqnarray}
is called the evidence lower bound (ELBO) and $\log p(\{\mathrm{x}^{(v)}\})$ is log-likelihood.

Minimizing KL divergence is equivalent to maximizing the ELBO.
Often $q(\mathrm{z},c|\{\mathrm{x}^{(v)}\})$ is assumed to be a mean-field distribution and can be readily factorized by
\begin{equation}
q(\mathrm{z},c|\{\mathrm{x}^{(v)}\})=q(\mathrm{z}|\{\mathrm{x}^{(v)}\})q(c|\{\mathrm{x}^{(v)}\}). \label{eqn:qx_cz}
\end{equation}

Due to the powerfulness of DNN to approximate non-linear function, we here introduce a neural network $g( \cdot )$ to infer $q(\mathrm{z}|\{\mathrm{x}^{(v)}\})$, with parameters $\{\phi^{(v)}\}^m_{v=1}$. That is, DNN is utilized to encode observed view data into latent representation. Meanwhile, to incorporate multi-view information, we propose a combined variational approximation $q(\mathrm{z}|\{\mathrm{x}^{(v)}\})$.
Considering the
importance of different views, we introduce a weight vector ${\rm w}=[w_{1},w_{2},...,w_{m}]^T$ ($w_{v}\geq 0, \sum w_{v} =1$ ) to fuse the distribution of hidden variables, so that 
the consistency and complementary of multi-view data can be better exploited.
In particular, we assume the variational approximation to the posterior of latent representation $\rm z$ to be a Gaussian by integrating information from multiple views as follows.
\begin{eqnarray}
[\tilde{\mu}^{(v)};\log(\tilde{\sigma}^{(v)})^2]&=&g(\mathrm{x}^{(v)};\phi^{(v)}), \label{eqn:mu_sig}\\
\tilde{\mu}&=&[\tilde{\mu}^{(1)},...,\tilde{\mu}^{(m)}] {\mathrm w}, \label{eqn:mu}\\
\tilde{\sigma}^{2}&=&[(\tilde{\sigma}^{(1)})^{2},...,(\tilde{\sigma}^{(m)})^{2}] {\rm w}, \label{eqn:sig}\\
q(\mathrm{z}|\{\mathrm{x}^{(v)}\})&=&\mathcal{N}(\mathrm{z}|\tilde{\mu},\tilde{\sigma}^{2}\mathbf{I}), \label{eqn:qx_z}
\end{eqnarray}
where $\mathbf{I}$ is an identity matrix with suitable dimension. In the standard VAE, each pair of $\tilde{\mu}^{(v)}$ and $(\tilde{\sigma}^{(v)})^2$ defines a Gaussian for latent variable $\rm z$ in the $v$-th view. We have fused the information in Eqs.~\eqref{eqn:mu_sig} - \eqref{eqn:qx_z}.

Furthermore, ELBO can be rewritten by,
\begin{flalign}
&\mathcal{L}_{\mathrm{E}\mathrm{L}\mathrm{B}\mathrm{O}}(\{\mathrm{x}^{(v)}\}) \nonumber \\ &=E_{q(\mathrm{z},c|\{\mathrm{x}^{(v)}\})}\left[\log\frac{p(\{\mathrm{x}^{(v)}\},\mathrm{z},c)}{q(\mathrm{z},c|\{\mathrm{x}^{(v)}\})}\right] \nonumber\\
&=\displaystyle \int_{\mathrm{z}}q(\mathrm{z}|\{\mathrm{x}^{(v)}\})\log\frac{p(\{\mathrm{x}^{(v)}\}|\mathrm{z})p(\mathrm{z})}{q(\mathrm{z}|\{\mathrm{x}^{(v)}\})}d\mathrm{z}\nonumber \\
&\ \ \ -\int_{\mathrm{z}}q(\mathrm{z}|\{\mathrm{x}^{(v)}\})D_{KL}(q(c|\{\mathrm{x}^{(v)}\})||p(c|\mathrm{z}))d\mathrm{z}.
\label{eqn:elbo2}
\end{flalign}
Hence, we set $D_{KL}(q(c|\{\mathrm{x}^{(v)}\})||p(c|\mathrm{z})) \equiv 0$ to maximize $\mathcal{L}_{\mathrm{E}\mathrm{L}\mathrm{B}\mathrm{O}}(\{\mathrm{x}^{(v)}\})$, due to the first term has no relationship with $c$ and the second term is non-negative. As a result, we use the following equation to compute $q(c|\{\mathrm{x}^{(v)}\})$, i.e.,
\begin{equation}
q(c|\{\mathrm{x}^{(v)}\})=p(c|\mathrm{z})\equiv\frac{p(c)p(\mathrm{z}|c)}{\sum_{c}p(c)p(\mathrm{z}|c)}.
 \label{eqn:qxv_c}
\end{equation}

This means we are proposing a mixture model for the latent prior $p(\rm z)$. Particularly we implement the latent prior $p(\rm z)$ as a Gaussian mixture as follows,
\begin{eqnarray}
p(c) &=& Cat(c|\pi), \label{eqn:pc} \\
p(\mathrm{z}|c)&=&\mathcal{N}(\mathrm{z}|\mu_{c},\sigma_{c}^{2}\mathbf{I}), \label{eqn:pc_z} \end{eqnarray}
where $Cat(c|\pi)$ is the categorical distribution with parameter $\pi = (\pi_1, ..., \pi_K)\in \mathbb{R}_{+}^{K}, ~\sum \pi_{c} =1$ such that $\pi_{c}$ ($c=1, ..., K$) is the prior probability for cluster $c$, and both $\mu_{c}$ and $\sigma_{c}^{2}$  ($c=1, ..., K$)  are the mean and the variance of the $c$-th Gaussian component, respectively.

Once the latent variable $\rm z$ is produced according to the GMM prior, the multi-view data generative process will defined as, for the binary observed data,
\begin{eqnarray}
\mu_{\theta^{(v)}}  &=& f(\mathrm{z};\theta^{(v)}), \\
p(\mathrm{x}^{(v)}|\mathrm{z})&=& \textup{Ber}(\mathrm{x}^{(v)} | \mu_{\theta^{(v)}}),
\label{eqn:pz_x}
\end{eqnarray}
where $f(\mathrm{z};\theta^{(v)})$ is a deep neural network whose input is $\mathrm{z}$ parameterized by $\theta^{(v)}$, $\textup{Ber}(\mu_{\theta^{(v)}})$ is multivariate Bernoulli distribution parameterized by $\mu_{\theta^{(v)}}$. Or for the continuous data,
\begin{align}
\mu_{\theta^{(v)}}  &= f_1(\mathrm{z};\theta^{(v)}), \label{eqn:u_theta}\\
\log(\sigma^2_{\theta^{(v)}})  &= f_2(\mathrm{z};\theta^{(v)}), \label{eqn:v_theta}\\
p(\mathrm{x}^{(v)}|\mathrm{z})&= \mathcal{N}(\mathrm{x}^{(v)} | \mu_{\theta^{(v)}}, \sigma^2_{\theta^{(v)}}),
\label{eqn:pz_x}
\end{align}
where $f_1(\cdot)$ and $f_2(\cdot)$ are all deep neural networks with appropriate parameters $\theta^{(v)}$, producing the mean and variance for the Gaussian likelihoods.   The generative process is depicted in the right part of Figure~\ref{fig:model}.

For the $v$-th view, since $\mathrm{x}^{(v)}$ and $c$ ~are independent conditioned on $\mathrm{z}$,  the joint probability $p(\{\mathrm{x}^{(v)}\}, \mathrm{z},c)$ can be decomposed by,
\begin{align}
\begin{aligned}
p(\{\mathrm{x}^{(v)}\},\mathrm{z},c)=&p(\{\mathrm{x}^{(v)}\}|\mathrm{z})p(\mathrm{z},c)\\
=&p(\{\mathrm{x}^{(v)}\}|\mathrm{z})p(\mathrm{z}|c)p(c)
\end{aligned}\label{eqn:pxvc}
\end{align}

Next, by using the \textit{reparameterization} trick and the Stochastic Gradient Variational Bayes (SGVB) \cite{KingmaWelling2014}, the objective function of our method for binary data can be formulated by,
\begin{flalign}
&\mathcal{L}_{\mathrm{E}\mathrm{L}\mathrm{B}\mathrm{O}}(\{\mathrm{x}^{(v)}\})= \nonumber \\
&\frac{1}{L}\sum_{v=1}^{m}\sum_{\iota=1}^{L}\sum_{i=1}^{D}\mathrm{x}_{i}^{(v)}\log\mu_{\theta^{(v)}}|_{i}^{l}+(1-\mathrm{x}_{i}^{(v)})\log(1-\mu_{\theta^{(v)}}|_{i}^{l}) \nonumber \\
&\ \ \ -\frac{1}{2}\sum_{c=1}^{K}\gamma_{c}\sum_{j=1}^{J}(\log\sigma_{c}^{2}|_{j}+\frac{\tilde{\sigma}^{2}|_{j}}{\sigma_{c}^{2}|_{j}}+\frac{(\tilde{\mu}|_{j}-\mu_{c}|_{j})^{2}}{\sigma_{c}^{2}|_{j}}) \nonumber \\
&\ \ \ +\displaystyle \sum_{c=1}^{K}\gamma_{c}\log\frac{\pi_{c}}{\gamma_{c}}+\frac{1}{2}\sum_{j=1}^{J}(1+\log\tilde{\sigma}^{2}|_{j})  \label{eqn:lebo3}
\end{flalign}
where $\mu_{\theta^{(v)}}$ is outputs of the DNN $f(\cdot)$, $L$ denotes the number of Monte Carlo samples in the SGVB estimator and is usually set to be 1. The dimension for $\mathrm{x}^{(v)}$ and $\mu_{\theta^{(v)}}$ is $D$ while the dimension for $\mu_{c}, \tilde{\mu}, \sigma_{c}^{2}$ and $\tilde{\sigma}^{2}$ is $J$. $\mathrm{x}_{i}^{(v)}$ denotes the $i$-th element of $\mathrm{x}^{(v)}$, $*|_{i}^{l}$ represents the $l$-th sample in the $i$-th element of $*$, and $*|_{j}$ means the $j$-th element of $*$. $\gamma_{c}$ denotes $q(c|\{\mathrm{x}^{(v)}\})$ for simplicity.

For the continuous data, the objective function is rewritten as:
\begin{flalign}
&\mathcal{L}_{\mathrm{E}\mathrm{L}\mathrm{B}\mathrm{O}}(\{\mathrm{x}^{(v)}\})= \nonumber \\
&\frac{1}{L}\sum_{v=1}^{m}\sum_{\iota=1}^{L}\sum_{i=1}^{D}-\frac{1}{2}\log2\pi\sigma^2_{\theta^{(v)}}|_{i}^{l}-\frac{(\mathrm{x}^{(v)}_{i}-\mu_{\theta^{(v)}}|_{i}^{l})^{2}}{2\sigma^2_{\theta^{(v)}}|_{i}^{l}}\nonumber \\
&\ \ \ -\frac{1}{2}\sum_{c=1}^{K}\gamma_{c}\sum_{j=1}^{J}(\log\sigma_{c}^{2}|_{j}+\frac{\tilde{\sigma}^{2}|_{j}}{\sigma_{c}^{2}|_{j}}+\frac{(\tilde{\mu}|_{j}-\mu_{c}|_{j})^{2}}{\sigma_{c}^{2}|_{j}}) \nonumber \\
&\ \ \ +\displaystyle \sum_{c=1}^{K}\gamma_{c}\log\frac{\pi_{c}}{\gamma_{c}}+\frac{1}{2}\sum_{j=1}^{J}(1+\log\tilde{\sigma}^{2}|_{j}) 
\label{eqn:lebo4}
\end{flalign}
where $\mu_{\theta^{(v)}}$ and $\sigma^2_{\theta^{(v)}}$ can be obtained by Eq. \eqref{eqn:u_theta} and Eq. \eqref{eqn:v_theta}, respectively. Intuitively, the first term of Eq. \eqref{eqn:lebo4} is used for reconstruction, and the rest is the KL divergence from the Gaussian mixture prior $p(\mathrm{z},c)$ to the variational posterior $q(\mathrm{z},c|\{\mathrm{x}^{(v)}\})$. As such, the model can not only generate the samples well, but make variational inference close to our hypothesis.

Note that although our model is also equipped with VAE and GMM, it is distinct from the existing work \cite{DuDuHe2017,JiangZhengTanTangZhou2017}. Our model focuses on multi-view clustering task by simultaneously learning the generative network, inference network and the weight of each view.

By a direct application of the chain rule and estimators, similar to the work \cite{DuDuHe2017,JiangZhengTanTangZhou2017}, the gradients of the loss for Eqs. \eqref{eqn:lebo3} and \eqref{eqn:lebo4} are calculated readily. To train the model, the estimated gradients in conjunction with standard stochastic gradient based optimization methods, such as SGD or Adam, are applied. Overall, using the mixed Gaussian latent variables, the proposed model can be trained by back-propagation with {\it reparameterization} trick. 
After training, the shared latent representation $\rm z$ is achieved for each sample $\mathrm{x}_i (i=1,2,...,n)$. Finally the final cluster assignment is computed by Eq. \eqref{eqn:qxv_c}.

\section{Experimental Results}
\subsection{Datasets}
To evaluate the performance of the proposed DMVCVAE, we select four real-world datasets including digits, object and facial images. A summary of the dataset statistics is also provided in Table~\ref{table:datasets}.
\begin{itemize}
\item[$\bullet$] \textbf{UCI digits}\footnote{
\url{https://archive.ics.uci.edu/ml/datasets/Multiple+Features}} consists of features of handwritten digits of 0 to 9 extracted from UCI machine learning repository \cite{DuaGraff2017}. It contains 2000 data points with 200 samples for each digit. 
These digits are represented by six types of features, including pixel averages in $2 \times 3$ windows (PIX) of dimension 240, Fourier coefficients of dimension 76, profile correlations (FAC) of dimension 216, Zernike moments (ZER) of dimension 47, Karhunen-Loeve coefficients (KAR) of dimension 64 and morphological features (MOR) of dimension 6.
\item[$\bullet$] \textbf{Caltech 101} is an object recognition dataset \cite{LiFergusPerona2004} containing 8677 images 
of 101 categories. We chose 7 classes of Caltech 101 with 1474 images, i.e., Face, Motorbikes, Dolla-Bill, Garfield, Snoopy, Stop-Sign and Windsor-Chair. 
There are six different views, including Gabor features of dimension of 48, wavelet moments of dimension 40, CENTRIST features of dimension 254, histogram of oriented gradients(HOG) of dimension 1984, GIST features of dimension 512, and local binary patterns (LBP) of dimension 928.
\item[$\bullet$] \textbf{ORL} contains 10 different images from each of 40 distinct subjects. For some subjects, the images were taken at different times with varying lighting, facial expressions and facial details. It consists of three types of features: intensity of dimension 4096, LBP features of dimension 3304 and Gabor features of dimension 6750.
\item[$\bullet$] \textbf{NUS-WIDE-Object (NUS)} is a dataset for object recognition which consists of 30000 images in 31 classes. We use 5 features provided by the web-site, i.e. 65 dimension color Histogram (CH), 226 dimension color moments (CM), 145 dimension color correlation
(CORR), 74 dimension edge distribution and 129 wavelet texture.
\end{itemize}
\setlength{\tabcolsep}{4pt}
\begin{table}
\begin{center}
\caption{Dataset Summary}
\label{table:datasets}
\begin{tabular}{c|c|c|c}
\hline
Datasets & \# of samples & \# of views & \# of classes\\
\hline
UCI digits  & 2,000 & 6 & 10\\
Caltech-7 & 1,474 & 6 & 7\\
ORL & 400 & 3 & 40\\
NUS-WIDE-Object & 30,000 & 5 & 31\\
\hline
\end{tabular}
\end{center}
\end{table}
\setlength{\tabcolsep}{1.4pt}

\subsection{Experiment Settings}
In our experiments, the fully connected network and same architecture settings as DEC \cite{XieGirshickFarhadi2016} are used. More specifically, the architectures of $g(\mathrm{x}^{(v)};\phi^{(v)})$ and $f(\mathrm{z};\theta^{(v)})$ are $d_v$-500-500-200-10 and 10-2000-500-500-$d_v$, respectively, where $d_v$ is input dimensionality of each view. We use Adam optimizer \cite{KingmaBa2014} to maximize the objective function, and set the learning rate to be 0.0001 with a decay of 0.9 for every 10 epochs.

Initializing the parameters of the deep neural network is usually utilized to avoid the problem that the model might get stuck in a undesirable local minima or saddle points. Here, we use layer-wise pre-training method \cite{bengio2007greedy} for training DNN $g(\cdot)$ and $f(\cdot)$. After pre-training, the network $g(\cdot)$ is adopted to project input data points into the latent representation $\rm z$, and then we perform $K$-means to $\rm z$ to obtain $K$ initial centroids of GMM ${\mu}_c ( c \in \lbrace 1,\cdots, K\rbrace$). Besides, the weights $\rm w$ of Eqs. \eqref{eqn:mu} and \eqref{eqn:sig} are initialized to $\frac1m$ for each view and the parameter of GMM $\pi_{k}$ is initialized to $\frac1K$.

Three popular metrics are used to evaluate the clustering performance, i.e. clustering accuracy (ACC), normalized mutual information (NMI) and adjusted rand index (ARI), in which the clustering accuracy is defined by
\begin{equation}
\textup{ACC}=\max_{m\in \mathcal{M}}\frac{\sum_{i=1}^N{1}\{l_i=m(c_i)\}}{N},\nonumber
\end{equation}
where $l_i$ is the ground-truth label, $c_i$ is the cluster assignment obtained by the model,
and $\mathcal{M}$ ranges over all possible one-to-one mappings between cluster assignment and labels. The mapping $m(\cdot)$ can be efficiently fulfilled by the Kuhn-Munkres algorithm \cite{ChenDonohoSaunders2001}.
NMI indicates the correlation between predicted labels and ground truth labels. ARI scales from $-1$ to 1, which measures the similarity between two data clusterings, higher value usually means better clustering performance. As each measure penalizes or favors different properties in the clustering, we report results on all the measures for a comprehensive evaluation.

\subsection{Baseline Algorithms}
We compare the proposed DMVCVAE with the following clustering methods including both shallow models and deep models.
\begin{itemize}
\item[$\bullet$] \textsl{Single View}: Choosing the single view of the best clustering performance using the graph Laplacian derived from and performing spectral clustering on it.
\item[$\bullet$] \textsl{Feature Concatenation} (abbreviated to Feature Concat.): Concatenating the features of all views and conducting spectral clustering on it.
\item[$\bullet$] \textsl{Kernel Addition}: Building an affinity matrix from every feature and taking an average of them, then inputting to a spectral clustering algorithm.
\item[$\bullet$] \textsl{MultiNMF}\cite{LiuWangGaoHan2013}: \textsl{Multi-view NMF} applies NMF to project each view data to the common latent subspace. This method can be roughly considered as one-layer version of our proposed method.
\item[$\bullet$] \textsl{LT-MSC}\cite{ZhangFuLiuLiuCao2015}: \textsl{Low-rank tensor constrained multi-view subspace clustering} proposes a multi-view clustering by considering the subspace representation matrices of different views as a tensor.
\item[$\bullet$] \textsl{SCMV-3DT}\cite{YinGaoXieGuo2018}: \textsl{Low-rank multi-view clustering in third-order tensor space via t-linear combination} using \textit{t-product} based on the circular convolution to reconstruct multi-view tensorial data by itself with sparse and low-rank penalty.
\item[$\bullet$] \textsl{DCCA} \cite{andrewAroraBilmesLivescu2013}: Providing flexible nonlinear representations with respect to the correlation objective measured on unseen data.
\item[$\bullet$] \textsl{DCCAE} \cite{WangAroraLivescuBilmes2015}: Combining the DCCA objective and reconstruction errors of the two views.
\item[$\bullet$] \textsl{VCCAP} \cite{wang2016deep}: Using a deep generative method to achieve a natural idea that the multiple views can be generated from a small set of shared latent variables.
\end{itemize}
In our experiments, $K$-means is utilized for six shallow methods to obtain the final clustering results. For three deep methods, DCCA, DCCAE and VCCAP, we use spectral clustering to perform the clustering, similar to the work \cite{WangAroraLivescuBilmes2015}.

\subsection{Performance Evaluation}
We first compare our method with six shallow models on the chosen test datasets. The parameter settings for the compared methods are done according to their authors' suggestions for their best clustering scores. The clustering performance of different methods are achieved by running 10 trials and reporting the average score of the performance measures, shown in Table~\ref{table:shallow}. The bold numbers highlight the best results.

As can be seen, except for the \textsl{Single View}, the other methods exploit all of views data with an improved performance than using a single view. In terms of all of these evaluation criteria, our proposed method consistently outperforms the shallow models for UCI digits and Caltech-7 datasets. In particularly, for Caltech-7, our method outperforms the second best algorithm in terms of ACC and NMI by 17.7\% and 25.0\%, respectively. While for ORL dataset, LT-MSC and SCMV-3DT achieves the best result in terms of NMI and ARI, respectively. This may be explained by the small size of ORL dataset, since large-scale datasets often lead to better performance for deep models. The results also verify that our model DMVCVAE significantly benefits from deep learning.

To further verify the performance of our approach among the deep models, we report the comparisons between the deep models, given in Tabel ~\ref{table:deep}. Since these three models can only handle two views data, we tested all the two view combination and the
best clustering score is reported finally. Specifically, FAC and KAR features are chosen in UCI digits, GIST and LBP features for Caltech-7, and LBP and Gabor features for ORL. For fair comparison, we perform the proposed model on the same views. From Tabel ~\ref{table:deep}, it is observed that our proposed method significantly outperforms others on all criteria.
\renewcommand{\arraystretch}{1.5}
\begin{table}
  \centering
  \fontsize{6.5}{8}\selectfont
  \caption{Clustering performance comparison between the propose model and shallows methods.}
  \label{table:shallow}
    \begin{tabular}{|c|c|c|c|c|c|c|c|c|c|}
    \hline
    \multirow{2}{*}{Methods}&
    \multicolumn{3}{c|}{UCI-digits}&\multicolumn{3}{c|}{Caltech-7}&\multicolumn{3}{c|}{ORL}\cr\cline{2-10}
    &ACC&NMI&ARI&ACC&NMI&ARI&ACC&NMI&ARI\cr
    \hline  
    Single View&0.6956&0.6424&0.7301&0.4100&0.4119&0.2582&0.6700&0.8477&0.5676\cr\hline
    Feature Concat. &0.6973&0.6973&0.6064&0.3800&0.3410&0.2048&0.6700&0.8329&0.5590\cr\hline
    Kernel Addition&0.7700&0.7456&0.3700&0.3936&0.2573&0.6570&0.6000&0.8062&0.4797\cr\hline
    \hline
    MultiNMF&0.7760&0.7041&0.6031&0.3602&0.3156&0.1965&0.6825&0.8393&0.5736\cr\hline
    LT-MSC&0.8422&0.8217&0.7584&0.5665&0.5914&0.4182&0.7587&{\bf 0.9094}&0.7093\cr\hline
    SCMV-3DT&0.9300&0.8608&0.8459&0.6246&0.6031&0.4693&0.7947&0.9088&{\bf 0.7381}\cr\hline
    \hline
    Ours&{\bf 0.9570}&{\bf 0.9166}&{\bf 0.9107}&{\bf 0.8014}&{\bf 0.8538}&{\bf 0.7048}&{\bf 0.7975}& 0.9013&0.7254\cr
    \hline
    \end{tabular}
\end{table}

\renewcommand{\arraystretch}{1.5}
\begin{table}
  \centering
  \fontsize{6.5}{8}\selectfont
  \caption{Clustering performance comparison among the deep models.}
  \label{table:deep}
    \begin{tabular}{|c|c|c|c|c|c|c|c|c|c|}
    \hline
    \multirow{2}{*}{Methods}&
    \multicolumn{3}{c|}{UCI-digits}&\multicolumn{3}{c|}{Caltech-7}&\multicolumn{3}{c|}{ORL}\cr\cline{2-10}
    &ACC&NMI&ARI&ACC&NMI&ARI&ACC&NMI&ARI\cr
    \hline 
    DCCA&0.8195&0.8020&0.7424&0.8242&0.6781&0.7131&0.6125&0.8094&0.4699\cr\hline
    DCCAE&0.8205&0.8057&0.7458&0.8462&0.7054&0.7319&0.6425&0,8115&0.5048\cr\hline
    VCCAP&0.7480&0.7320&0.6277&0.8372&0.6301&0.7206&0.4150&0.6440&0.2418\cr\hline
    Ours&{\bf 0.8875}&{\bf 0.8076}&{\bf 0.7765}&{\bf 0.8568}&{\bf 0.7386}&{\bf 0.7826}&{\bf 0.6950}&{\bf 0.8356}&{\bf 0.5643}\cr
    \hline
    \end{tabular}
\end{table}
\subsection{Visualizations}
\begin{figure*}[htbp]
\centering
\subfigure[DCCAE]{\begin{minipage}[t]{0.3\linewidth}
\centering
\includegraphics[width=4cm]{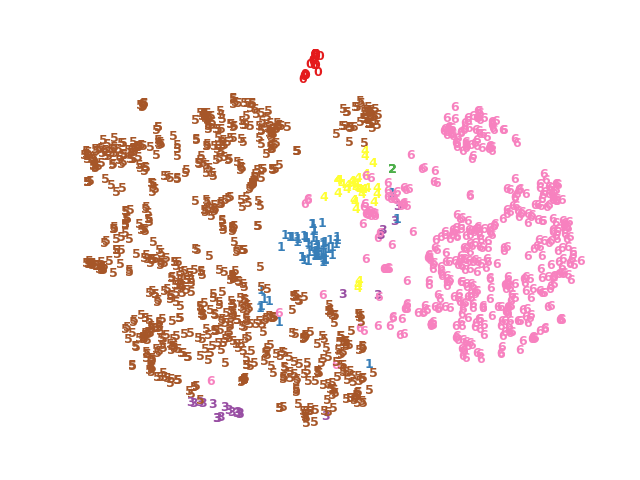}
\end{minipage}
}
\subfigure[VCCAP]{\begin{minipage}[t]{0.3\linewidth}
\centering
\includegraphics[width=4cm]{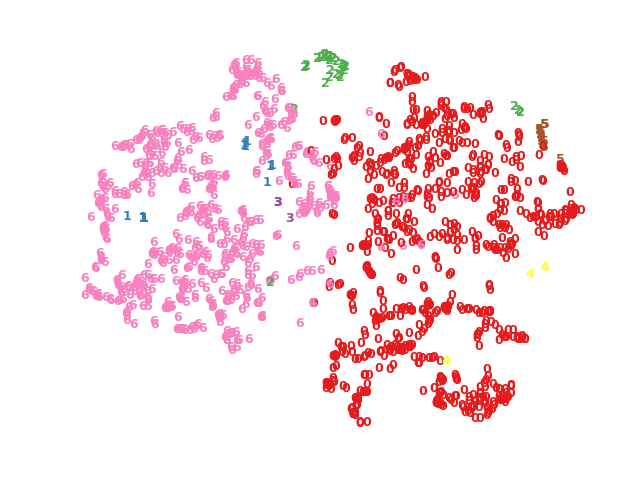}
\end{minipage}
}
\subfigure[DMVCVAE]{\begin{minipage}[t]{0.3\linewidth}
\centering
\includegraphics[width=4cm]{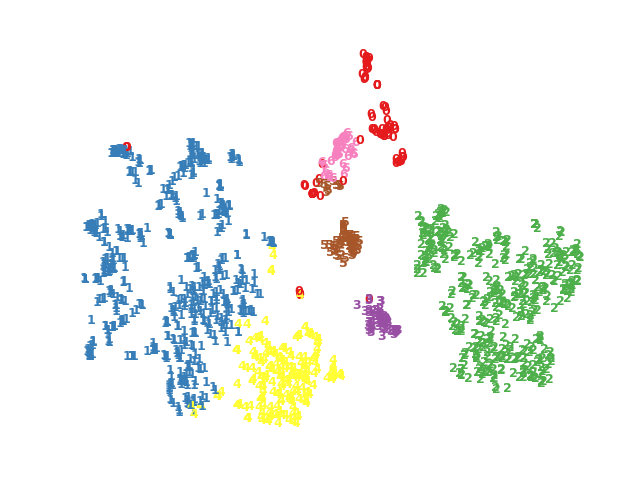}
\end{minipage}
}
\centering
\caption{Visualization to show the latent subspaces of Caltech-7 dataset.}
\label{fig:vis_all}
\end{figure*}

\begin{figure*}[htbp]
\centering
\subfigure[Epoch 10]{
\includegraphics[width= 3.56cm]{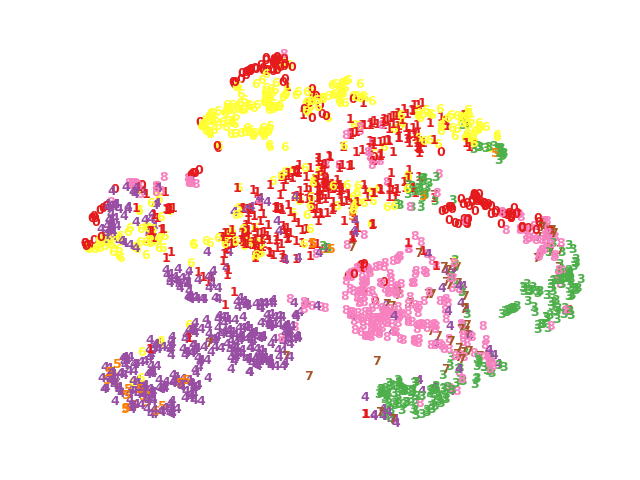}
}
\subfigure[Epoch 40]{
\includegraphics[width= 3.56cm]{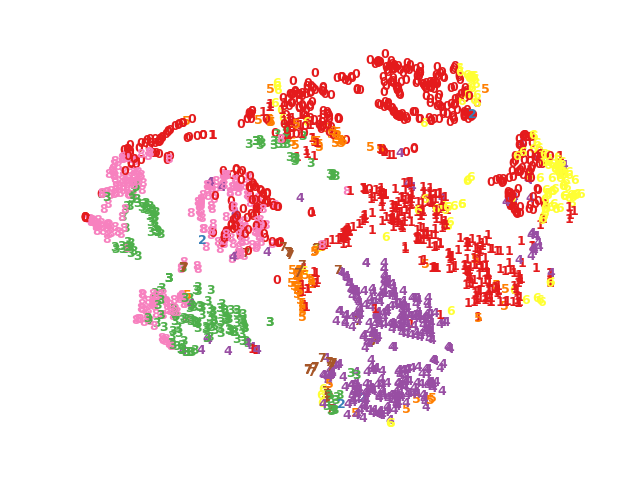}
}
\subfigure[Epoch 70]{
\includegraphics[width= 3.56cm]{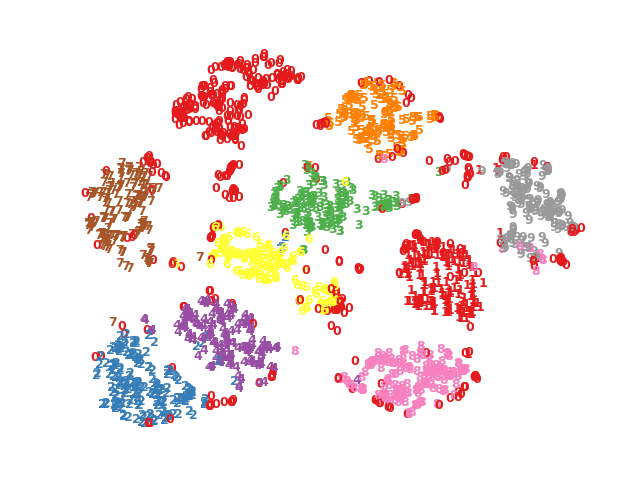}
}
\subfigure[Epoch 100]{
\includegraphics[width= 3.56cm]{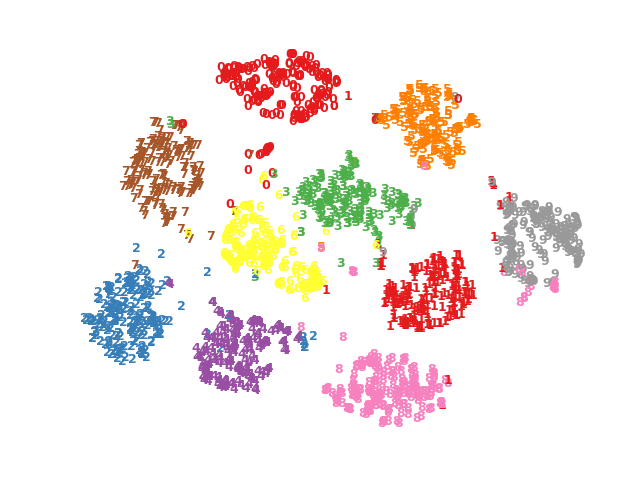}
}
\caption{Visualization to show the latent subspaces of UCI digits by DMVCVAE visualization from epoch 10 to 100.}
\label{fig:vis_epoch}
\end{figure*}

In Figure~\ref{fig:vis_all}, we 
visualize the latent space on Caltech-7 dataset by various deep models. t-SNE \cite{maaten2008visualizing} is applied to reducing the dimensionality to 2-dimensional space. It can be observed that the embedding learned by DMVCVAE is better than that by DCCAE and VCCAP. Figure~\ref{fig:vis_epoch} shows the learned representations of DMVCVAE on UCI digits dataset.  Specifically, we see that, as training progressing, the latent feature clusters become more and more separated, suggesting that the overall architecture motivates seeking informative representations with better clustering performance.

\subsection {Experiment on large-scale multi-view data}
With the unprecedentedly explosive growth in the volume of visual data, how to effectively segment large-scale multi-view data becomes an interesting but challenging problem \cite{LiNieHuangHuang2015,ZhangLiuShenShenShao2018}. Therefore, we further test our model on the large-scale dataset, i.e., NUS-WIDE-Object. As the aforementioned compared methods cannot handle the large-scale data, we compare with the recent work, such as Large-Scale Multi-View Spectral Clustering (LSMVSC) \cite{LiNieHuangHuang2015} and Binary Multi-View Clustering (BMVC) \cite{ZhangLiuShenShenShao2018}. In this experiment, we replace the ARI measure with PURITY such that the comparison will be fair\footnote{Here we cited the reported results from their original papers as the lack of the corresponding source codes. $``-"$ means there is no report in the original paper.}. By the similar settings, the clustering results are reported in Table~\ref{table:Largescale}. As can be seen, our proposed approach achieved better clustering performance against the compared ones and verified the strong capacity on handling large-scale multi-view clustering.
\begin{table}
\centering
\fontsize{6.5}{8}\selectfont
\caption{Clustering performance for large-scale dataset. }
\label{table:Largescale}
\begin{tabular}{|c|c|c|c|}
\hline
\multirow{2}{*}{Methods}&
\multicolumn{3}{c|}{NUS-WIDE-Object }\cr\cline{2-4}
&ACC&NMI&PURITY\cr
\hline
LSMVSC& -- &0.1493&0.2821\cr\hline
BMVC&0.1680&0.1621&0.2872\cr\hline
Ours& \textbf{0.1909} & \textbf{0.2129} & \textbf{0.3168}\cr
\hline
\end{tabular}
\end{table}

\section{Conclusions}
In this paper, we proposed a novel multi-view clustering algorithm by learning a shared latent representation under the VAE framework. The shared latent embeddings, multi-view weights and deep autoencoders networks are simultaneously learned in a unified framework such that the final clustering assignment is intuitively achieved. Experimental results show that the proposed method can provide better clustering solutions than other state-of-the-art approaches, including the shallow models and deep models.

{\small
\bibliographystyle{ieee}

}

\end{document}